\documentclass[11pt,a4paper]{article}
\usepackage{times,latexsym}
\usepackage{url}
\usepackage[T1]{fontenc}
\usepackage{graphicx}
\usepackage{booktabs}
\usepackage{comment}
\usepackage{amssymb}
\usepackage{enumitem}
\usepackage{array,multirow}
\usepackage{soul}
\usepackage{color, colortbl}
\usepackage[normalem]{ulem}
\usepackage{makecell}

\usepackage{arabtex}
\usepackage{tipa}
\usepackage[T1]{fontenc}
\usepackage[utf8]{inputenc}

    \usepackage[acceptedWithA]{tacl2018v2}
\usepackage[]{tacl2018v2}

\providecommand\ac[1]{\textcolor{teal}{AC: \textit{#1}}}

\providecommand{\siamak}[1]{{\small \color{orange} [SS: #1]}}
\providecommand{\daniel}[1]{{\small \color{red} [DK: #1]}}
\providecommand{\changed}[1]{#1}

\providecommand{\pedram}[1]{{\small \color{blue} [\textbf{PH}: #1]}}

\newcommand{\FB}[1]{\textcolor{purple}{Faeze: #1}}

\usepackage{xspace,mfirstuc,tabulary}

\newif\iftaclinstructions
\taclinstructionsfalse 
\iftaclinstructions
\renewcommand{\confidential}{}
\renewcommand{\anonsubtext}{(No author info supplied here, for consistency with
TACL-submission anonymization requirements)}
\newcommand{\instr}
\fi

\iftaclpubformat 

\else

\fi

\newcommand{\parsiglue}{\textsc{ParsiNLU}}

\title{
\parsiglue: A Suite of High-level NLP Tasks for Persian}
\title{\parsiglue: A Suite of High-level NLP Challenges for Persian}
\title{
    \vspace*{-0.5in}
    {{\small \hfill TACL'21}\\
    \vspace*{.25in}} 
\parsiglue: \\ A Suite of Language Understanding Challenges for Persian}

\author{
\fontsize{11pt}{11pt}\selectfont
 \makecell{Daniel Khashabi$^{1}$ \hfill \; Arman Cohan$^1$ \hfill \; Siamak Shakeri$^{2}$ \hfill \; Pedram Hosseini$^{3}$ \hfill \; Pouya Pezeshkpour$^4$  \\
 Malihe Alikhani$^5$ \hfill Moin Aminnaseri$^6$ \hfill Marzieh Bitaab$^7$ \hfill Faeze Brahman$^{8}$  \hfill\\ 
 Sarik Ghazarian$^9$ \hfill Mozhdeh Gheini$^9$ \hfill Arman Kabiri$^{10}$ \hfill Rabeeh Karimi Mahabadi$^{11}$ \\ 
 Omid Memarrast$^{12}$ \hfill Ahmadreza Mosallanezhad$^{7}$ \hfill Erfan Noury$^{13}$ \hfill Shahab Raji$^{14}$ \\ 
Mohammad Sadegh Rasooli$^{15}$ \hfill Sepideh Sadeghi$^{2}$ \hfill Erfan Sadeqi Azer$^{2}$ \hfill Niloofar Safi Samghabadi$^{16}$ \\ Mahsa Shafaei$^{17}$ \hfill Saber Sheybani$^{18}$ \hfill Ali Tazarv$^{4}$  \hfill Yadollah Yaghoobzadeh$^{19}$} 
\\
\fontsize{8pt}{8pt}\selectfont
\makecell{ 
$^{1}$Allen Institute for AI, $^{2}$Google, $^{3}$George Washington U., $^{4}$UC Irvine, $^{5}$U. of Pittsburgh, $^{6}$TaskRabbit, $^{7}$Arizona State U., $^{8}$UC Santa Cruz\\
$^{9}$U. of Southern California, $^{10}$IMRSV Data Labs, $^{11}$EPFL, $^{12}$U. of Illinois - Chicago, $^{13}$U. of Maryland Baltimore County\\  $^{14}$Rutgers U., $^{15}$U. of Pennsylvania, $^{16}$Expedia Inc., $^{17}$U. of Houston, $^{18}$Indiana U. - Bloomington, $^{19}$Microsoft
}
}

\newcommand\blfootnote[1]{%
  \begingroup
  \renewcommand\thefootnote{}\footnote{#1}%
  \addtocounter{footnote}{-1}%
  \endgroup
}

\date{}

\begin{document}
\maketitle
\begin{abstract}
\blfootnote{$\star$ The point of view of the authors are their own and not attributable to the company they work for.}

Despite the progress made in recent years in addressing  natural language understanding (NLU) challenges, the majority of this progress remains to be concentrated on resource-rich languages like English. 
 This work focuses on Persian language, one of the widely spoken languages in the world, and yet there are few NLU datasets available for this language.
 The availability of high-quality evaluation datasets is a necessity for reliable assessment of the progress on different NLU tasks and domains. 
  We introduce \parsiglue{}, the first benchmark in Persian language that includes a range of language understanding tasks --- \emph{Reading Comprehension}, \emph{Textual Entailment}, etc.
  These datasets are collected in a multitude of ways, often involving manual annotations by native speakers. 
  This results in over 14.5$k$ new instances across 
  6 distinct NLU tasks.
  Besides, we present the first results on state-of-the-art monolingual and multi-lingual pre-trained language models on this benchmark and compare them with human performance, which provides valuable insights into our ability to tackle natural language understanding challenges in Persian. 
  We hope \parsiglue{} fosters further research and advances in Persian language understanding.\footnote{
    \url{https://git.io/JIuRO}
    \label{github-link}
  }
\end{abstract}


\section{Introduction}

In recent years, considerable progress has been made in building stronger NLU models, particularly supported by high-quality benchmarks~\cite{bowman2015large,rajpurkar2016squad,wang2019superglue} for resourceful languages like English. However, in many other languages, such benchmarks remain scarce, unfortunately, stagnating the progress towards language understanding in these languages.

In this work, we focus on developing NLU benchmarks for Persian \changed{(also known as ``Farsi'')}.
This language has many attributes that 
make it distinct from other well-studied languages. 
In terms of script, Persian is similar to Semitic languages (e.g., Arabic).
Linguistically, however, Persian is an Indo-European language~\cite{masica1993indo} and thus distantly related to most of the languages of Europe as well as the northern part of the Indian subcontinent.
Such attributes make Persian a unique case to study in terms of language technologies. 
Although Persian is 
\changed{a widely spoken language}
~\cite{simons2017ethnologue}, our ability to evaluate performance and measure the progress of NLU models on this language remains limited. This is mainly due to the lack of major language understanding benchmarks that can evaluate progress on a diverse range of tasks.

\begin{figure*}
    \centering
    \includegraphics[scale=0.61,trim=0.37cm 0.6cm 0cm 0.7cm]{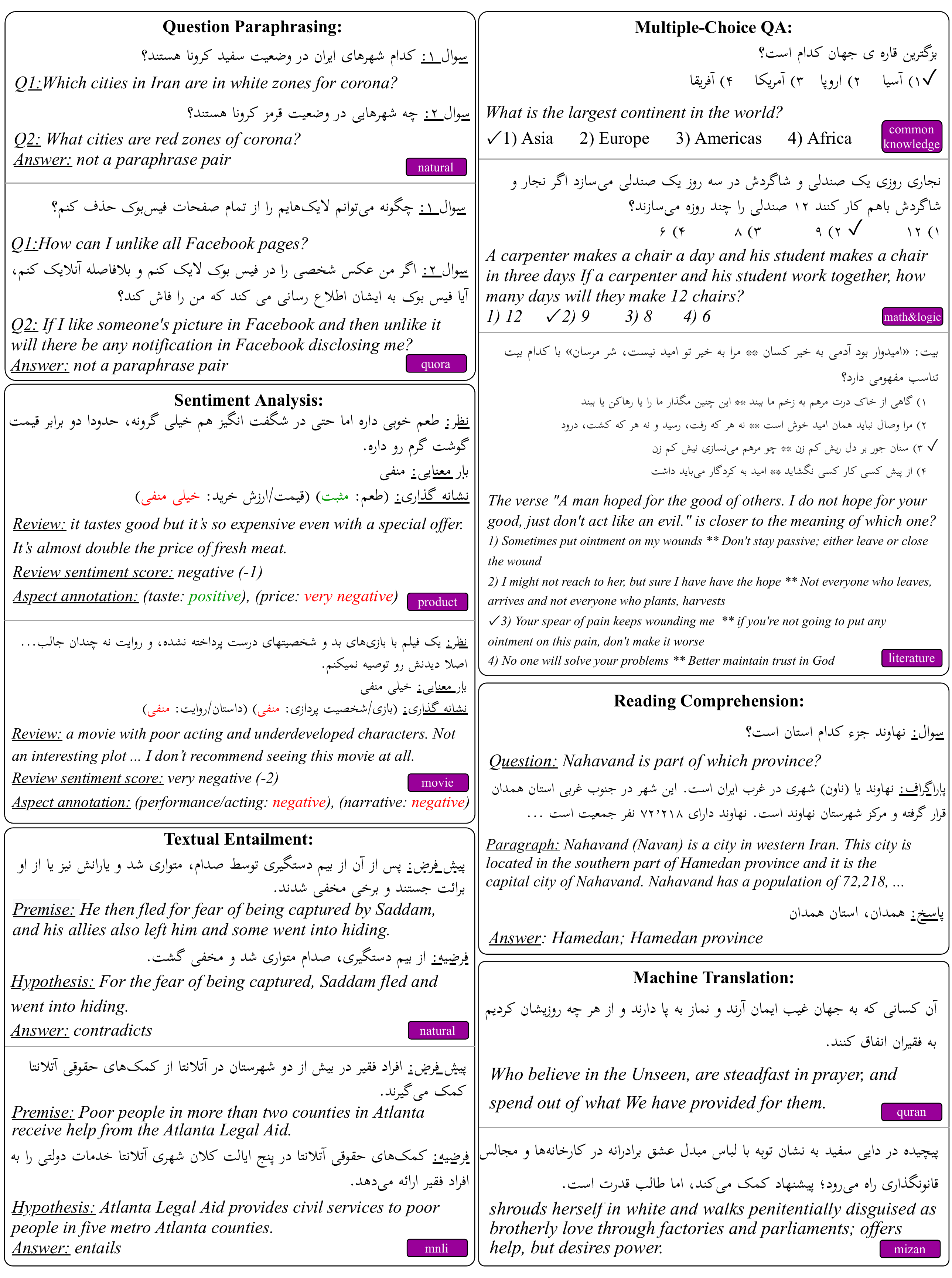}
    \caption{Examples of the \parsiglue\ tasks. For each task (other than Machine Translation \changed{which already contains English phrases) we show the English translations for ease of communication to non-Persian readers.} The purple tags indicate the example category, according to their construction (explained in the main text under Section~\ref{subsec:constructing:tasks}).} 
    \label{fig:examples_figure}
\end{figure*}

In this work, we present \parsiglue{}, a collection of NLU challenges for Persian.\footnote{
    We focus on the standard Iranian Persian, spoken by over 80 million people.
    There are other dialects of Persian spoken in other countries, e.g., Afghanistan and Tajikistan. 
} 
\parsiglue{} contains challenges for \emph{reading comprehension}, \emph{multiple-choice question-answering}, 
\emph{textual entailment}, 
\emph{sentiment analysis}, 
\emph{question paraphrasing}, and \emph{machine translation} (examples in Fig.~\ref{fig:examples_figure}). 
\parsiglue{} offers data for tasks that have never been explored before \changed{in the context of the Persian language}. 
We are not aware of any publicly available dataset for Persian \emph{question answering} (\S\ref{subsec:multiple-choice}), \emph{reading comprehension} (\S\ref{subsec:reading:comprehension}), and  \emph{paraphrasing} (\S\ref{subsec:qqp}). 
For the rest of the tasks, we improve at least one aspect of the existing datasets (e.g., better data construction, more comprehensive evaluation, and evaluation of less investigated genres or domains).  
To ensure the quality of the presented challenge tasks, 
we rely on the annotations from native Persian speakers or novel data collection techniques, such as, search engine auto-complete~(\S\ref{subsec:reading:comprehension}), and past collegiate exams~(\S\ref{subsec:multiple-choice}).
To the best of our knowledge, this is the first comprehensive collection of its own, composed of a variety of Persian NLU tasks. 


We conduct a collection of empirical work (\S\ref{sec:experiments}) to establish 
the difficulty of \parsiglue. 
We benchmark each \parsiglue\ task via collecting  
state-of-the-art multi-lingual and mono-lingual 
LMs, as well as estimating the human upper bound scores. 
The gap between human and machine baselines indicate the need for further research and stronger models
for Persian. 
We hope that the release of \parsiglue\ 
will encourage more research on Persian NLP. 

\section{Related Work}
\label{sec:related:work}

\paragraph{Cross-lingual benchmarks.}
There are several recent cross-lingual benchmarks; however, almost none includes Persian: 
XNLI~\cite{conneau2018xnli} for entailment, 
PWNS-X~\cite{yang2019paws} for paraphrasing, 
XCOPA~\cite{ponti2020xcopa} for choice of plausible alternatives, 
XQuAD, MLQA, TyDI and MKQA~\cite{artetxe2019cross,lewis2019mlqa,clark2020tydi,longpre2020mkqa} for reading comprehension. 
These datasets have also been integrated as part of multi-task multilingual evaluation suites such as the XTREME~\cite{hu2020xtreme} and  
XGLUE~\cite{liang2020xglue}. 
Unfortunately, the Persian portion of the former benchmark covers only two tagging tasks (POS and NER) and the latter does not 
\changed{cover Persian}.

\paragraph{NLU benchmarks for other languages.}

Benchmarks like GLUE~\cite{wang2019superglue} encourage  development of better and stronger models on a diverse set of challenges. There have been several efforts to create GLUE-like benchmarks for other languages; for example, 
CLUE for Chinese~\cite{xu2020clue},
GLUECoS for Hindi~\cite{2020gluecos}, 
and RussianSuperGLUE~\cite{shavrina2020russiansuperglue}. 
We view \parsiglue{} in the same family of benchmarks, dedicated to the Persian language.

\paragraph{NLU Datasets for Persian.}
Prior work on creating evaluation resources for the Persian language 
\changed{has focused on low-level tasks in narrow domains}
(e.g., datasets for POS~\cite{bijankhan2004role}, NER~\cite{shahshahani2019payma}, Parsing~\cite{seraji2013uppsala}).
\changed{Complementary to these efforts, }
we aim at providing an 
NLU evaluation benchmark for Persian, consisting of a wide variety of tasks. Below we mention several related works and how we build upon them. 

\changed{
FarsTail~\cite{amirkhani2020farstail} is a concurrent work on the  \emph{entailment} task, where the dataset is constructed semi-automatically based on existing multiple-choice exams. 
Different from this work, our entailment datasets are built with the annotations of native speakers of Persian and some use of machine translation (\S\ref{subsec:entailment}). 
Therefore, we hypothesize our construction represents a slightly different distribution than that of FarsTail. 
}

\changed{
There is a rich set of works on Persian \emph{sentiment analysis}. 
We build upon these works and differ from them in the following manners: 
(a) 
The existing work mainly focuses on 
\emph{document-level} sentiment identification 
which does not capture
the nuanced judgements with respect to 
aspects and entities of
the context~\cite[inter alia]{hosseinzadehbendarkheili2019product,sharami2020deepsentipers}. 
In addition to such \emph{document-level} annotations,
we provide \emph{aspect-level} sentiment annotations (\S\ref{subsec:sa}).
(b) 
The majority of existing resources, such as MirasOpinion~\cite{ashrafi-asli-etal-2020-optimizing} focus on 
binary or ternary sentiment classes.
However, our annotations contain a more granular sentiment intensity with five labels~(\S\ref{subsec:sa}).
(c) 
Compared to the aspect-level datasets~\citet{hosseini2018sentipers,ataei2019pars}, 
we cover two relatively less investigated domains:  \textit{food \& beverages} and \textit{movies}, each posing new challenges for Persian sentiment analysis. 
}

\emph{Machine translation} of Persian $\rightleftarrows$ English is one of the few tasks that has enjoyed decent attention~\cite{2004opus,mohaghegh-etal-2010-improved,pilevar2011tep,mohaghegh-etal-2011-improving,rasooli-etal-2013-orthographic,karimi2017extracting,kashefi2018mizan,khojasteh2020lscp}. 
Unfortunately, most published work for this task focus on niche domains and datasets. 
    Our contribution to this task is compiling a set of high-quality  evaluation sets from a broad range of domains,
    \changed{
        based on the existing datasets as well as datasets introduced in this work. 
    }
    The hope is that this will help future work on Persian MT to evaluate their systems on a variety of domains to get a more realistic measure of machine translation. 
    
    
To the best of our knowledge, this is the first work that publishes an evaluation benchmark for Persian language, promoting future studies on several NLU tasks such as \emph{question answering} (\S\ref{subsec:multiple-choice}), \emph{reading comprehension} (\S\ref{subsec:reading:comprehension}), and \emph{paraphrasing} (\S\ref{subsec:qqp}), among others.

\section{\parsiglue}
\label{sec:parsi:glue}

\subsection{Design Considerations}
\label{subsec:desiderata}
We now discuss possible design choices for constructing the dataset and the underlying reasons.

\paragraph{Naturally-occurring instances. }
A common way of collecting data for low-resource languages has been using automated translation of the benchmark datasets of high-resource languages~\cite{artetxe2019cross,ponti2020xcopa}. 
This can be a poor practice as recent investigations have shown translation artifacts in data gathered via translation of existing tasks~\cite{artetxe2020translation}. 
It is important for any NLP dataset to reflect the \emph{natural} distribution of the target language tokens and their associated cultural contexts. 
Therefore, one should \emph{avoid} over-reliance on automatic conversion of resources from high-resource languages
to minimize any unnatural instances or artifacts~\cite{khvalchik2020departamento}.

\paragraph{Experts, over crowdworkers. }
While crowdsourcing has been the common approach for building datasets, we choose to work with few native Persian speakers to construct the dataset. 
Crowdworkers are difficult to train and often generate more noisy annotations. However, expert annotators that are closely familiar with the task at hand often generate better quality annotations. 
Using crowdworkers is further complicated by the fact that crowdsourcing platforms do not have an active community of Persian-speaking workers due to limited international financial transactions and 
crowdsourcing platforms.
A study done by \citet[Table 6]{pavlick2014language} shows that 
there are almost no crowd-workers for Persian on the Amazon Mechanical Turk platform. 

\subsection{Constructing \parsiglue\ tasks}
\label{subsec:constructing:tasks}
Examples are shown in Fig.~\ref{fig:examples_figure}.
We now explain the data construction of each task.

\subsubsection{Reading Comprehension}
\label{subsec:reading:comprehension}
We use the commonly used definition of reading-comprehension task: 
\changed{
    extracting a substring from a given context \emph{paragraph} that answers a given \emph{question}. 
}

SQuAD~\cite{rajpurkar2016squad} is one of the most popular reading comprehension datasets in English. Similar datasets to SQuAD are developed in other languages using varying degrees of human or semi-automatic translation techniques:  KorQuAD for Korean~\cite{lim2019korquad1}, MMQA for Hindi~\cite{gupta2018mmqa}, etc. 
For constructing our reading comprehension tasks, we avoid using SQuAD as a source and employ a process resembling that of \citet{kwiatkowski2019natural} that would lead to more natural questions. 

\paragraph{Collecting questions.}
Our efforts to translate questions from the English dataset indicated that such questions are often about topics that are not of much importance in Persian. For instance, there are many questions in SQuAD~\cite{rajpurkar2016squad} about major US sports events (e.g., Superbowl, NFL)
or western civilization history that 
might not be common among Persian speakers. 
Instead, we follow a pipeline that is more similar to the one introduced by~\citet{kwiatkowski2019natural}, setting our goal to annotate answers for an existing naturalistic set of questions in Persian, as opposed to writing questions for existing paragraphs.

Unlike \citet{kwiatkowski2019natural}, we do not have direct access to query logs. Thus we follow the approach of \citet{berant2013semantic,khashabi2021gooaq} which relies on a query auto-completion API for collecting questions. 
Similarly, we use Google's auto-completion\footnote{http://google.com/complete/search?client=chrome\&q=...} which enables us to mine a rich, yet a natural set of questions in Persian as it is reflective of popular questions posed by users of Google.

We start with a seed set of question terms (e.g., 
{\small``\setfarsi\novocalize \<^che kesI>''} [\textipa{che kasI}] meaning ``who'', and 
{\small``\setfarsi\novocalize \<kojA>''} [\textipa{kojA}] meaning ``where'')
We bootstrap based on this set, by repeatedly querying parts of previously-extracted questions, in order to discover a longer and richer set of questions.
We hypothesize that such questions extracted from the auto-complete algorithm, 
are highly reflective of popular questions posed by Persian-speaking users of Google. 
We filter out any results shorter than 5 tokens as they are often incomplete questions. 
This process yields over 50$k$ questions. 

Subsequently, we automatically filter out open-ended questions with no concrete answers (e.g., 
{\small``\setfarsi\novocalize \<natIjeh\nospace ye bAzI bA ^zApon?>''} [\textipa{n\ae tIdZe ye bAzI bA ZApon?}] meaning ``What is the results of the game with Japan?'').
Our filtering was guided by the observation that typically more complete questions lead to Google results that include  well-established sources (such as Wikipedia). 
Hence, we perform this filtering by retrieving the Google search results\footnote{https://github.com/MarioVilas/googlesearch} for each question and checking if any 
of the top 10 search results overlap with a pre-defined list of credible websites.\footnote{fa.wikipedia.org, bbcpersian.com, etc.}
We keep only the questions that match this criterion. 

\paragraph{Annotating paragraphs and answers.}
In this step, native speakers of Persian select a paragraph and an answer span within the paragraph that answers each of the questions.
At the first step, the annotators read the question and correct any grammatical errors and typos (e.g., 
{\small``\setfarsi\novocalize \<otsAn>''} [\textipa{otsAn}] 
is corrected to 
{\small``\setfarsi\novocalize \<ostAn>''} [\textipa{ostAn}] ``state''). 
Next, they annotate all \emph{the \changed{minimal and} coherent spans} that contains the answer to the question, from a paragraph obtained from a relevant web page
(from the Google search results retrieved from an earlier step). 
Whenever possible, we annotate all valid spans as the answer \changed {(for example, {\small``\setfarsi\novocalize \<hamedAn>''} [\textipa{hæmedAn}] and {\small``\setfarsi\novocalize \<ostAn hamedAn>''} [\textipa{ostAn e hæmedAn}], as shown in Fig.~\ref{fig:examples_figure})}. 
The paragraph that contains this answer is also annotated as the context of the question. 

Overall, 6 native-speaker annotators annotated a collection of 1.3$k$ question-answer-paragraph triplets (Table~\ref{tab:statistics}).

\paragraph{Annotation quality.}
To 
\changed{ensure}
the quality of the annotations, 
\changed{
  the answers to each question were labeled by two independent annotators.
  Any misalignment of the answer spans or missing any valid spans were indicated as disagreements. 
}
Such disagreements were resolved in further adjudication.

\subsubsection{Multiple-Choice QA}
\label{subsec:multiple-choice}
Multiple-choice questions  
are one of the common formats for evaluation of fact-retrieval and reasoning~\cite{richardson2013mctest,clark2019f}. 
Following prior works, we define the task as: given a natural language question, pick the correct answer among a list of multiple candidates.
A key difference from reading comprehension (\S\ref{subsec:reading:comprehension}) is that the instances are open-domain (i.e., 
no context paragraph is provided).
Hence, a model would either need to retrieve external supporting documents 
or have stored the necessary knowledge internally to be able to answer the question. 

\paragraph{Sources of questions.}
We use existing sources of multiple-choice questions, rather than annotating new ones. We collect the questions from a variety of sources: 
(i) The literature questions of the annual college entrance exams in Iran, for the past 15 years. These questions often involve the understanding of poetry and their implied meaning, knowledge of Persian grammar, and the history of literature. 
(ii) Employment exams that are expected to assess an individual's depth in various topics (accounting, teaching, mathematics, logic, etc).
(iii) Common knowledge questions, which involve questions about topics such as basic science, history, or geography.

Most of the above sources are scanned copies of the original exams in image format. 
We use an existing Persian OCR tool to convert the image data to a textual format.\footnote{
    https://www.sobhe.ir/alefba/
}
Then 4 annotators fix any mistakes made by the OCR system and convert the result into a structured format. 
Overall, this yields \textit{2460} questions with an average of \textit{4.0} candidate answers (Table~\ref{tab:statistics}). 
Additionally, the task comes with a label indicating the type of knowledge it requires: 
`literature' (understanding of literary expressions), 
`common-knowledge' (encyclopedic knowledge or everyday activities), and `math \& logic' (logical or mathematical problems). Examples from each category of questions are included in Fig.~\ref{fig:examples_figure}.

\paragraph{Annotation quality.}
To further examine the quality of the annotations, we randomly sampled 100 questions from the annotations and cross-checked the OCR output with the original data. We discovered that 94 of such questions exactly matched the original data, and the rest required minor modifications. 
\changed{We thus conclude that the annotated data has a high quality.}

\subsubsection{Aspect-Based Sentiment Analysis}
\label{subsec:sa}
Sentiment Analysis (SA) is the study of opinions (i.e., positive, negative, or neutral sentiment) expressed in a given text~\cite{liu2012sentiment}. 
Aspect-based Sentiment Analysis (ABSA) is a more fine-grained SA that aims to extract aspects of entities mentioned in the text and determine sentiment toward these aspects~\cite{pontiki-etal-2014-semeval}. 
For instance, \emph{``it tastes good but it's so expensive ...''} (Fig.~\ref{fig:examples_figure}) conveys \emph{positive} and \emph{negative} sentiments with respect to \emph{taste} and \emph{price} aspects of the mentioned product (entity), respectively.   


\paragraph{Annotation scheme.} 
We follow the existing ABSA scheme~\cite{pontiki-etal-2014-semeval}. For every review, we do two types of annotations: 
(1) we assign an overall sentiment to each review, selecting from one of the following values: \emph{very-negative, negative, neutral, positive, very positive,} and \emph{mixed}. 
The \emph{mixed} category indicates reviews 
where none of the sentiments are dominant (mix of positive and negative, or borderline cases), hence it is hard to detect the primary sentiment of a review. 
We also assign \textit{neutral} label to reviews that express no clear sentiment toward an entity or any aspect of 
it. 
(2) we annotate pairs of 
$(a, s)$
where, $a$ is an  \textit{aspect} 
that belongs to a predefined set of aspects for each domain 
and $s$ expresses the sentiment toward the \textit{aspect} $a$.

\paragraph{Collecting reviews.} 
At first, 
we
collect 
reviews from two different domains: (1) \emph{food \& beverages} and 2) \emph{movies}. We chose these domains since they are relatively less investigated 
in the existing literature (see \S\ref{sec:related:work} for the past work). 
For the \emph{food \& beverages} category, we extracted\footnote{https://github.com/rajabzz/digikala-crawler} reviews from the online grocery section of Digikala,\footnote{https://www.digikala.com/} and for the \emph{movie} reviews category, we crawled reviews from Tiwall.\footnote{https://www.tiwall.com/} Both of these websites are well-known and popular 
websites 
among Persian speakers.

\paragraph{Defining aspects.} 
Following ABSA scheme, we predefined a set of aspects for each domain. For \textit{food \& beverages}, 
we crawled Digikala and retrieved all listed aspects for product reviews in the \textit{food \& beverages} category. Subsequently, we manually aggregated the extracted aspects  
and merged those 
with significant semantic overlap.
We also added \textit{taste/smell} as a new aspect category since users frequently commented on this aspect. For \emph{movie} reviews, we created an initial list of aspects based on the movie review aspects defined by~\citet{thet2010aspect}. 
In consultation with a movie critic, 
we resolved the potential overlaps among aspect categories and created a set of aspects that capture various perspectives of movie reviews. Overall, this process resulted in 6 and 7 aspects for \textit{food \& beverages} and \textit{movie review} domains, respectively (Table~\ref{fig:sentiment_aspects}). 

\begin{table}[]
    \centering
    \includegraphics[scale=0.61,trim=7.7cm 15.1cm 5cm 1.5cm]{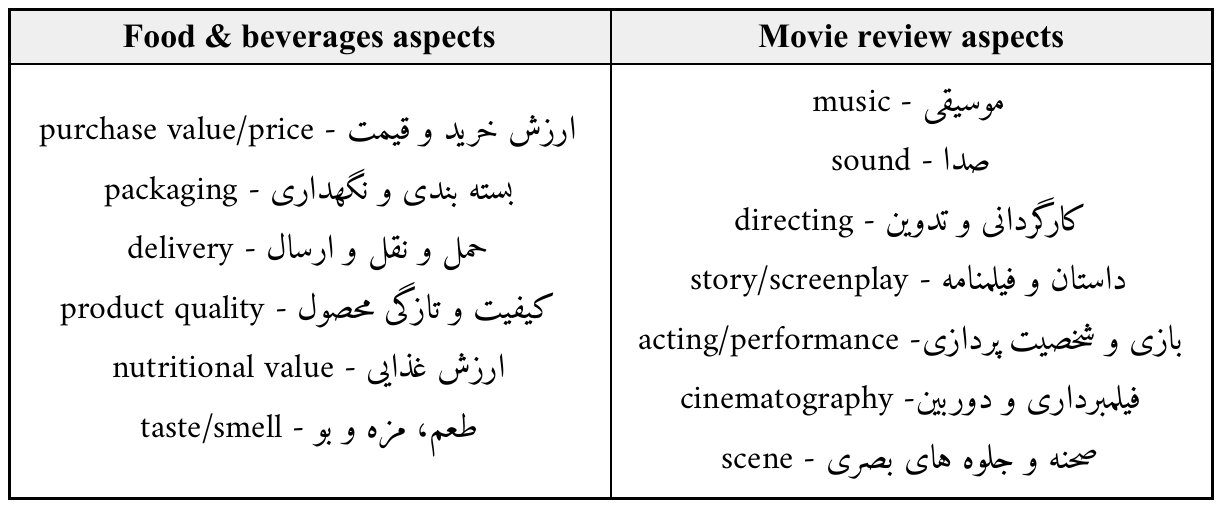}
    \caption{The predefined sentiment aspects (\S\ref{subsec:sa}). 
    }
    \label{fig:sentiment_aspects}
\end{table}

After defining the sentiment aspects, we trained four native speaker annotators 
for the final round of annotations. 
This results in 2423 instances for the sentiment task (Table~\ref{tab:statistics}).

\paragraph{Annotation  quality.} 
To measure the quality of the annotations, we randomly selected 100 samples from each domain and calculated the Inter-Annotator Agreement (IAA) using Cohen's kappa~\cite{cohen1960coefficient} on annotations elicited from two independent annotators. 
Based on the computed IAA values, there is a \textit{substantial} agreement on sub-task 1 (0.76), and \textit{moderate} agreement on sub-tasks 2 and 3 (0.49 and 0.47, resp.). 


\begin{table}[t]
    \centering
    \resizebox{.99\linewidth}{!}{
    \begin{tabular}{clc}
        \toprule
            Task & Attribute & Statistic \\
            \midrule
            \multirow{4}{*}{\rotatebox[origin=c]{90}{\parbox[c]{1.9cm}{\centering \small Reading Comprehension}}} 
            & \# of instances &  1300 \\ 
            & avg. question length (tokens) & 6.3 \\ 
            & avg. paragraph length (tokens) & 94.6 \\ 
            & avg. answer length (tokens) & 7.6 \\ 
        \midrule
            \multirow{5}{*}{\rotatebox[origin=c]{90}{\parbox[c]{1.9cm}{\centering \small Multiple-Choice QA}}} 
            & \# of instances  & 2460 \\ 
            & \% of `literature' questions  & 834 \\ 
            & \% of `common-knowledge' questions  & 949 \\ 
            & \% of `math \& logic' questions  & 677 \\ 
            & avg. \# of candidates & 4.0 \\ 
        \midrule
        \multirow{5}{*}{\rotatebox[origin=c]{90}{\parbox[c]{1.9cm}{\centering \small Sentiment Analysis}}} 
            & \# of instances & 2423 \\
            &  \% of `food \& beverages' reviews  & 1917  \\
            &  \% of `movie' reviews  & 506  \\
            &  avg. length of reviews (words) & 22.01 \\ 
            & \# of annotated pairs of (aspect, sentiment) & 2539 \\
        \midrule
        \multirow{5}{*}{\rotatebox[origin=c]{90}{\parbox[c]{1.9cm}{\centering \small Textual Entailment}}} 
             & \# of instances & 2,700 \\
             & \% of `natural' instances  & 1,370 \\ 
             & \% of `mnli' instances  & 1,330 \\ 
             & avg. length of premises (tokens)  & 23.4 \\ 
             & avg. length of hypotheses (tokens)  & 11.8 \\ 
        \midrule
        \multirow{5}{*}{\rotatebox[origin=c]{90}{\parbox[c]{1.9cm}{\centering \small Question Paraphrasing}}} 
            & \# of instances & 4,644 \\
             & \% of `natural' instances & 2,521 \\ 
             & \% of `qqp' instances  & 2,123 \\ 
             & avg. length of Q1 (tokens)  & 10.7 \\ 
             & avg. length of Q2 (tokens) & 11.0 \\ 
        \midrule
        \multirow{6}{*}{\rotatebox[origin=c]{90}{\parbox[c]{1.9cm}{\centering \small Machine Translation }}}
            & \# of instances & 47,745 \\
            & \% of `QP' subset & 489  \\
            & \% of `Quran' subset & 6,236  \\
            & \% of `Bible' subset & 31,020  \\
            & \% of `Mizan' subset (eval. only) & 10,000  \\
        \bottomrule
    \end{tabular}
    }
    \caption{Statistics on various subsets of the dataset.}
    \label{tab:statistics}
\end{table}

\changed{
\paragraph{Distribution of the labels.}
Here we report the distribution of the labels for this task.
Fig.~\ref{fig:sentiment:label:distribution} shows the distribution of the document-level sentiment labels. As expected, most reviews are associated with extreme sentiments (very positive or very negative) and a relatively small portion of them are neutral. There is also a non-negligible portion of the reviews that contains mixed sentiments (partially positive and partially negative). 
\begin{figure}[h]
    \centering
    \includegraphics[scale=0.41]{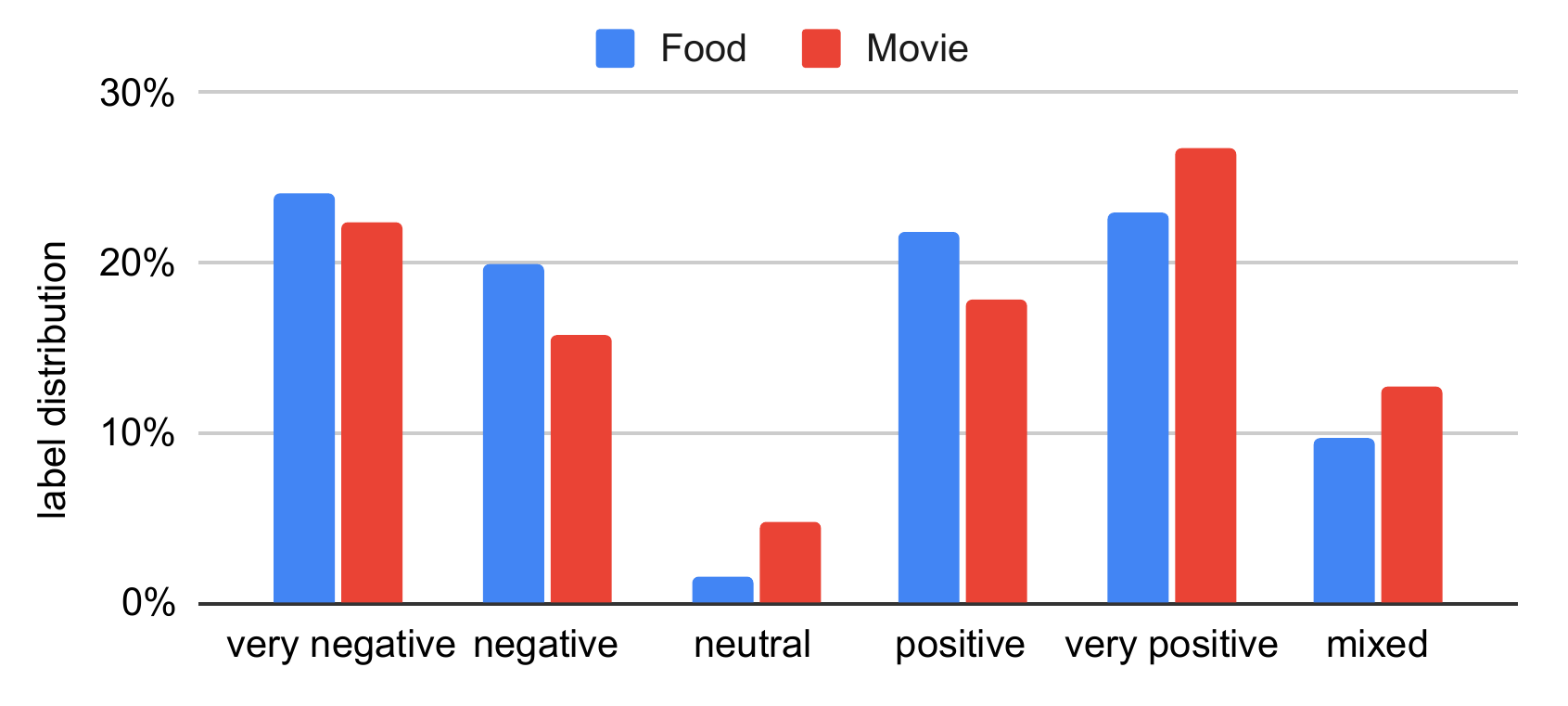}
    \caption{The distribution of the overall sentiments labels (document-level).}
    \label{fig:sentiment:label:distribution}
\end{figure}
}

\subsubsection{Textual Entailment}
\label{subsec:entailment}
Textual Entailment~\cite{2013Dagan,bowman2015large} 
is typically defined as a 3-way classification to determine whether a \emph{hypothesis} sentence \emph{entails}, \emph{contradicts}, or is \emph{neutral} with respect to a given \emph{premise} sentence.


We construct two subsets:
(i) based on available natural sentences, 
and (ii) based on the available English entailment dataset. 
The former approach yields high-quality instances,
however, it is a relatively slower annotation task. 
The latter is slightly easier, but yields less interesting instances.

\paragraph{Based on natural sentences.}
We start with randomly sampled raw sentences, selected from 3 different resources: Miras,\footnote{https://github.com/miras-tech/MirasText} 
Persian Wikipedia and VOA corpus.\footnote{https://jon.dehdari.org/corpora/}
In this random sampling process, we specifically sample sentences that contain
conjunctive adverbs 
(e.g, 
{\small``\setfarsi\novocalize \<ammA>''} [\textipa{amA}] meaning ``but''),
along with their preceding sentences. 
We chose such examples as there is a higher chance that these sentences naturally contain inference relationships. 
We ask annotators to consider both sentences and write a premise and  corresponding entailing, contradicting, and neutral sentences, whichever they deem appropriate. To minimize annotation artifacts and avoid creating an artificially easy dataset, we specifically instruct annotators to avoid using simple modifications, such as simply negating a sentence or changing a word to its synonym.
For the rest of the work, we refer to this set as the `natural' set. 

\paragraph{Based on existing datasets.}
In this approach, we use existing datasets in English. 
We start with MNLI dataset~\cite{williams2018broad} and translate them with the publicly available Google Translate API.\footnote{https://cloud.google.com/translate} 
Subsequently, expert annotators carefully review and fix inaccurate translations.  
Furthermore, each translated document is reviewed by a native-speaker annotator to correct the translational mistakes. 
Our annotations show that about $66.4\%$ of the translated documents have gone through some form of correction by our annotators. 
For the rest of the draft, we refer to this set as `mnli'. 

Overall, our two-pronged construction with 6 annotators results in 2.7$k$ entailment instances (Table~\ref{tab:statistics}). Examples from each collected subset are included in Fig.~\ref{fig:examples_figure}. 

\paragraph{Annotation quality.}
To verify the annotation quality, we quantify the agreement of 3 independent annotators, on 150 random examples. 
On this subset, we observe a Fleiss Kappa~\cite{fleiss1971measuring} of 
0.77, indicating a \emph{substantial} inter-annotator agreement~\cite{landis1977measurement}.

\changed{
\paragraph{Distribution of the labels.}
As the label distribution (Fig.~\ref{fig:label:dist:entailment}) shows, the distribution of the labels across the three categories are not far from uniform distribution. 

\begin{figure}[h]
    \centering
    \includegraphics[scale=0.57]{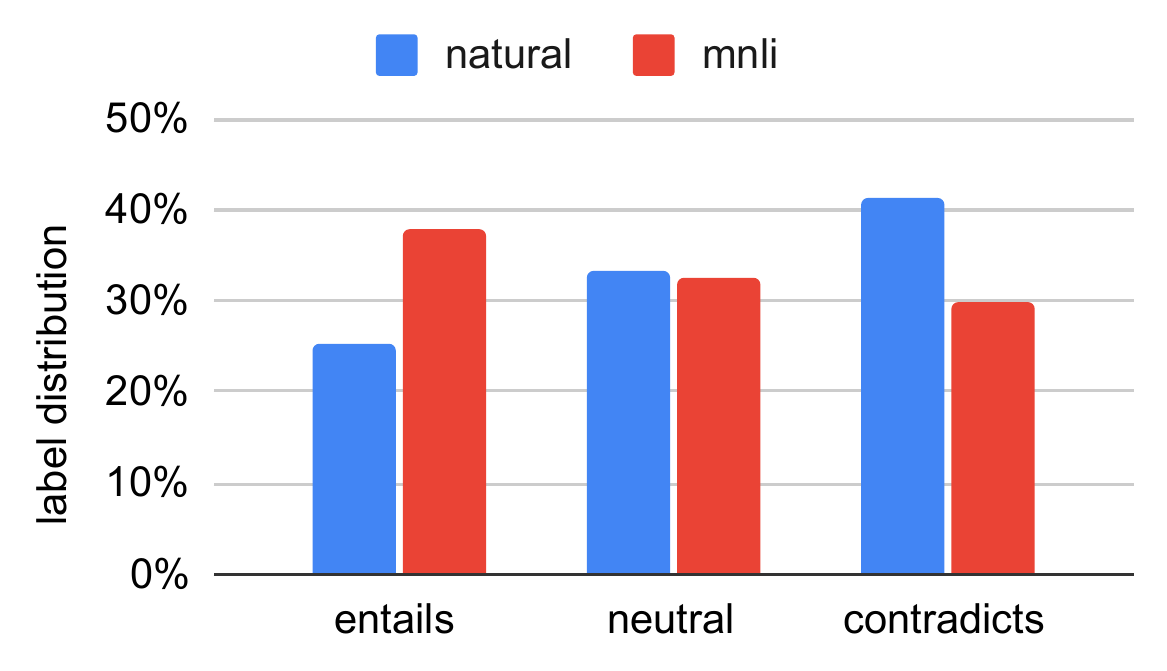}
    \caption{The distribution of the labels for the entailment task.}
    \label{fig:label:dist:entailment}
\end{figure}
}

\subsubsection{Question Paraphrasing}
\label{subsec:qqp}
This task is defined as determining whether two given questions are paraphrases or not. 
This task has been previously used to improve down-stream applications like document retrieval~\cite{zukerman2002lexical,callison2006improved,duboue2006answering}. 

Similar to the construction of the entailment 
task (\S\ref{subsec:entailment}), we take two different approaches:
(i) based on available natural sentences, (ii) based an existing English question paraphrasing dataset. 

\paragraph{Based on natural sentences.}
We start with questions mined using Google auto-complete (\S\ref{subsec:reading:comprehension}) as well as an additional set of questions mined from Persian discussion forums.\footnote{http://javabkoo.com/} We create pairs of questions with high token overlap. 
Each pair is annotated as \emph{paraphrase} or \emph{not-paraphrase} by native-speakers. 
We drop the pair if any of the questions is incomplete. 
For the rest of this document, we refer to this subset as `natural'.

\paragraph{Based on existing datasets.}
We start with the QQP dataset,\footnote{
https://www.kaggle.com/c/quora-question-pairs
\label{ft:qqp}
} which is a dataset of English question-pairs, and translate it with Google Translate API. 
Later, expert annotators carefully review the translations and amend any inaccuracies.
We observe that 
about $65.6\%$ of the translated documents have gone through some form of correction by our annotators. 

Overall, the annotations involved 4 annotators and resulted in 4682 question paraphrasing instances (Table~\ref{tab:statistics}).
Examples from each collected subset are included in Fig.~\ref{fig:examples_figure}. 

\paragraph{Annotation quality.}
\changed{
After the annotation of the earlier steps, the example were reviewed by another annotators familiar with the task. 
The disagreements were labeled and adjudicated among the annotators, in order to ensure the quality of the resulting labels.
}

\changed{
\paragraph{Distribution of the labels.}
As the label distribution shows (Fig.~\ref{fig:label:dist:qqp}), the label distributions of the two splits (`qqp' vs `natural') are not much different. 
\begin{figure}[h]
    \centering
    \includegraphics[scale=0.55]{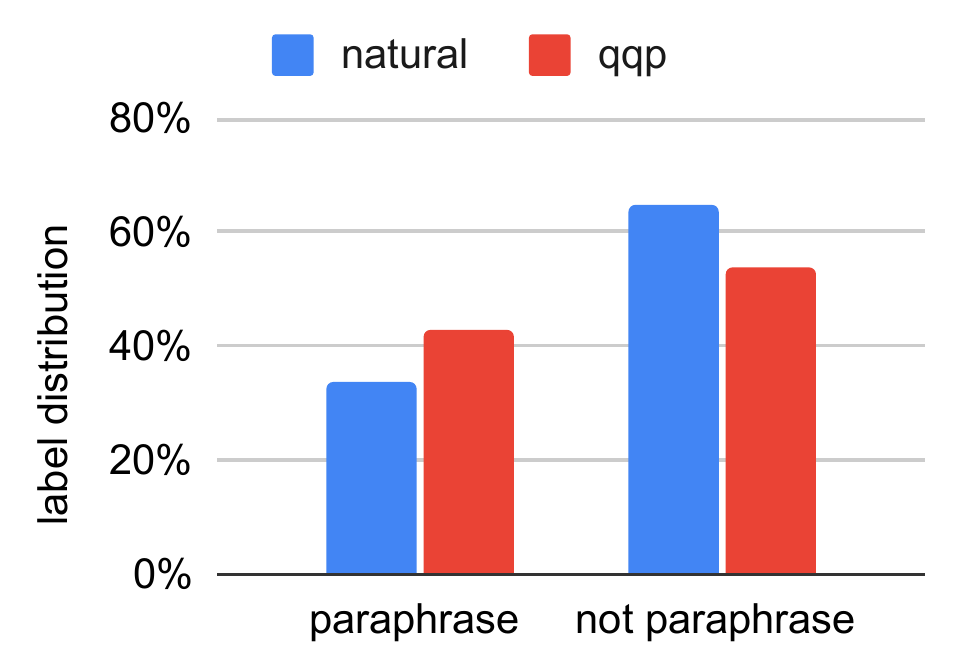}
    \caption{Label distribution for the query paraphrasing task.}
    \label{fig:label:dist:qqp}
\end{figure}

}

\begin{table}[]
    \centering
    \small
    \begin{tabular}{cccc}
        \toprule
        Task & Train & Dev & Eval  \\
        \midrule
        Reading Comprehension & 600 & 125 & 575  \\ 
        Multiple-Choice & 1271  & 139 & 1050  \\ 
        Sentiment Analysis & 1894  & 235 & 294 \\ 
        Textual Entailment & 756 & 271 & 1,751  \\ 
        Question Paraphrasing & 1,830 & 898 & 1,916 \\ 
        Machine Translation & $1.6m$ & $2k$ & $47k$ \\ 
        \bottomrule
    \end{tabular}
    \caption{Split sizes for different tasks.}
    \label{tab:split_sizes}
\end{table}

\definecolor{Gray}{gray}{0.9}

\begin{table*}[th]
    \centering
    \small
    \resizebox{\textwidth}{!}{
    \begin{tabular}{cccccccccc}
        \toprule
        \parbox[t]{2mm}{\multirow{2}{*}{\rotatebox[origin=c]{90}{\small Setup}}}  & Model $\downarrow$ - Task $\rightarrow$  &
        \parbox[c]{2.6cm}{\centering \small Reading Comprehension}
        & 
        \multicolumn{3}{c}{\parbox[c]{5.8cm}{\centering \small Multiple-Choice Question Answering}} 
        & 
        \multicolumn{2}{c}{\parbox[c]{3.6cm}{\centering \small Textual Entailment}} 
        & 
        \multicolumn{2}{c}{\parbox[c]{3.6cm}{\centering \small Question Paraphrasing}} 
        \\
        \cmidrule(lr){2-2}  \cmidrule(lr){3-3}  \cmidrule(lr){4-6} \cmidrule(lr){7-8} \cmidrule(lr){9-10} 
        
        &  Subtask $\rightarrow$  & 
        all & 
        literature & com-know & math \& logic & 
        natural & mnli & 
        natural & qqp \\
        \midrule
        \parbox[t]{2mm}{\multirow{7}{*}{\rotatebox[origin=c]{90}{\small trained on Persian}}} 
        & mBERT (base)    & 49.0 &  30.1.   & 28.7 & 33.8 & 48.7	& 51.6  & 80.4 & 75.3 \\
        & WikiBERT (base) & 39.2 & 36.9 &30.2  & 34.1 & 52.8 &  52.6	  & 80.0 & 75.5 \\ 
        & ParsBERT (base) & 40.7 & 33.4&28.6 &32.5  &   51.8 & 53.9 &79.4 & 72.0 \\ 
        & mT5 (small) & 30.9 &  33.7 & 23.7 & \bf{39.1}  & 51.9 & 51.0 & 75.2 & 72.0 \\ 
        & mT5 (base)  & 42.6 &   34.0 & 24.0 & 36.9   & 57.8 & 59.9 & 79.1 & 75.1 \\
        & mT5 (large) & 49.2 &   32.6 & 27.1 & 38.9  & 69.1 & 71.6 & 84.6 & 76.6 \\ 
        & mT5 (XL)    & 70.4 &    33.7 & 27.7 & 38.9   & {\bf 77.2} & 74.5 & 88.6 & 80.3 \\ 
        \midrule
        \parbox[t]{2mm}{\multirow{4}{*}{\rotatebox[origin=c]{90}{\small trained on} \rotatebox[origin=c]{90}{\small English}}} 
        & mT5 (small)  & 33.0 &    20.9  &	25.7  &	28.9      & 45.1	& 55.6 & 73.5 & 75.1 \\ 
        & mT5 (base)   & 53.4 &     23.4  &	23.4  &	24.3      & 44.4	& 43.3 & 83.2 & 81.8 \\ 
        & mT5 (large)  & 67.4 &     27.4  &	33.1  &	25.4    & 46.5	& 54.9 & 88.1 & 86.6  \\  
        & mT5 (XL)     & 68.2 &   28.3  &	\bf{38.6}  &	22.0   & 66.2	& 77.8 & {\bf 89.2} & {\bf 87.0} \\ 
        \midrule
        \parbox[t]{2mm}{\multirow{4}{*}{\rotatebox[origin=c]{90}{\small trained on} \rotatebox[origin=c]{90}{\small Per + Eng}}} 
        & mT5 (small) & 45.3 & 30.9	& 24.9	& 36.6 & 53.3 &	56.2 & 77.9 & 71.3 \\ 
        & mT5 (base)  & 63.9 & 32.3	& 24.0	& 37.7 & 57.8 &	63.9 & 80.2 & 73.4 \\ 
        & mT5 (large) & 73.6 & 30.6	& 28.9	& 38.6 & 70.9 & 72.5 & 85.3 & 78.9 \\  
        & mT5 (XL)    & {\bf 74.7} & \bf{38.0} & 33.7	& 38.0 & 75.5 &	{\bf 78.7} & 88.2 & 80.3\\ 
        \midrule
        & Human & 86.2 & 80.0 & 85.0 & 85.0 & 87.1 & 90.2 & 92.3 & 88.4 \\ 
        \bottomrule
    \end{tabular}
    }
    \resizebox{\textwidth}{!}{
    \begin{tabular}{cccccccccccccccccccccccc}
        \toprule
         \parbox[t]{2mm}{\multirow{2}{*}{\rotatebox[origin=r]{90}{\small Setup}}} & Model $\downarrow$ - Task $\rightarrow$ &
        \multicolumn{2}{c}{\parbox[c]{2.2cm}{\centering \small Sentiment (sentence sent.)}} 
        & 
        \multicolumn{2}{c}{\parbox[c]{2.0cm}{\centering \small Sentiment (aspect ext.)}} 
        & 
        \multicolumn{2}{c}{\parbox[c]{2.0cm}{\centering \small Sentiment (aspect sent.)}} 
        & 
        \multicolumn{4}{c}{\parbox[c]{5.3cm}{\centering \small Machine Translation (\changed{Eng} $\rightarrow$ \changed{Per})}}
        & 
        \multicolumn{4}{c}{\parbox[c]{5.3cm}{\centering \small Machine Translation (\changed{Per} $\rightarrow$ \changed{Eng})}}
        \\
        \cmidrule(lr){2-2}  \cmidrule(lr){3-4} \cmidrule(lr){5-6} \cmidrule(lr){7-8}  \cmidrule(lr){9-12} \cmidrule(lr){13-16}
        & Subtask $\rightarrow$  & 
        food & 
        movies & 
        food & 
        movies & 
        food & 
        movies & 
        quran & bible & qqp & mizan &
        quran & bible & qqp & mizan \\
        \midrule
        \parbox[t]{2mm}{\multirow{7}{*}{\rotatebox[origin=c]{90}{\small trained on our data}}} & mBERT (base) & 55.2 & 48.6 & 87.1 & 73.24 & 53.9 & 34.7 & - & - & - & - & - & - & - & - \\ 
        & WikiBERT (base) & 52.0 & 58.5 & 91.9 & 78.0 & 56.5 & 41.6 & - & - & - & - & - & - & - & - \\ 
        & ParsBERT (base) & 59.1 & 56.8 & 91.1 & 76.8 & 53.9 & 37.6 & - & - & - & - & - & - & - & - \\ 
        & mT5 (small) & 54.6 & 49.4 & 86.4 & 78.6 & 52.4 & 40.6 & 10.2 & 2.1 & 22.2 & 8.4 & 20.6 & 2.5 & 22.9 & 14.6  \\
        & mT5 (base) & 56.6 & 52.9 & 88.6 & 80.5 & 52.9 & 46.5 & 11.4 & 2.1 & 27.3 & 9.4 & 22.8 & 2.5 & 34.6 & 14.9 \\ 
        & mT5 (large) & 62.9  & { \bf 72.5 } & {\bf 92.2 } & 85.0 &  58.1  & 53.5 & 11.9 & 2.1 & {\bf 24.8} & 10.6 &  24.7 & 2.4 & 35.1 & 16.4 \\
        & mT5 (XL) &  {\bf 63.1 } & 70.6 &  92.0 & { \bf 85.8} & \bf{ 58.9 } & { \bf 54.5 } & {\bf 13.5 } & {\bf 2.2 } & 20.0  & {\bf 11.0 } & 30.0 & {\bf 2.6} & 33.7 & {\bf 19.3} \\ 
        \midrule
        \parbox[t]{2mm}{\multirow{4}{*}{\rotatebox[origin=c]{90}{\small trained on} \rotatebox[origin=c]{90}{\small English}}} 
        & mT5 (small) & - & -  & - & - & - & - & - & - & - & - & 6.6 & 1.9 & 7.7 & 3.7\\ 
        & mT5 (base) &  - & -  & - & - & - & -  & - & - & - & - &  11.5 & 2.1 & 14.0 & 5.7 \\ 
        & mT5 (large) & - & -  & - & - & - & - & - & - & - & - & 20.2 & 2.3 & 21.0 & 7.4 \\ 
        & mT5 (XL) & - & - & -  & - & - & - & - & - & - & -  & 25.6 & 2.3 &  30.7 & 9.7  \\ 
        \midrule
        \parbox[t]{2mm}{\multirow{4}{*}{\rotatebox[origin=c]{90}{\small trained on} \rotatebox[origin=c]{90}{\small Per + Eng}}} 
        & mT5 (small)  & - & -  & - & - & - & - & - & - & - & - & 19.2 & 2.5 & 25.6 & 12.1 \\ 
        & mT5 (base)   & - & -  & - & - & - & - & - & - & - & - & 24.1 & 2.4 & 36.0 & 14.8 \\ 
        & mT5 (large)  & - & -  & - & - & - & - & - & - & - & - & 29.9 & 2.6 & 36.5 & 18.1 \\ 
        & mT5 (XL)     & - & -  & - & - & - & - & - & - & - & - & {\bf 33.4} & 2.6 & {\bf 41.0 } & 18.2 \\ 
        \midrule
         & Human & 88.4 & 90.3 & 93.1 & 91.6 & 71.0 & 61.6 & - &- & - & -&  - & - & - & - \\
        \bottomrule
    \end{tabular}
    }
    
    \caption{
        Evaluation of \emph{Persian}-only models (top), \emph{English}-only (middle) and \emph{Persian+English} (bottom) models on Persian tasks.
        Best baseline scores are indicated as \textbf{bold}.
    }
    \label{tab:main:results:table}
\end{table*}

\subsubsection{Machine Translation}
\label{subsec:machine-translation}
We consider the task of translating
a given English sentence into Persian, and vice versa. 

This task is one of the few for which several resources are available in the literature~\cite{kashefi2018mizan,prokopidis2016parallel,pilevar2011tep}.
One major limitation is that there is no widely adopted comprehensive assessment of this task: most of the works are often limited to narrow domains, and the generalization across different styles of text is rarely studied. 
Our contribution is to put together a collection of evaluation sets, from various domains to encourage a more holistic \emph{evaluation} set. 

Our proposed evaluation sets consist of the followings: 
(i) \emph{Quran}: Quran has been translated into many languages, including English and Persian~\cite{2004opus}. We use several different translations of Quran to create high-quality evaluation sets (10 gold standard translations for each direction). Having multiple gold standards is particularly helpful for the automatic evaluation of machine translation since such metrics work best when provided with several gold standards~\cite{gupta2019investigating}. 
(ii)   \emph{Bible}: similarly, we use Persian and English versions of Bible\footnote{https://github.com/christos-c/bible-corpus} as another evaluation set.   
(iii) \emph{QQP}: using the data obtained in the construction of question paraphrasing task  (\S\ref{subsec:qqp}) to create an evaluation set for translating language questions.  
(iv) \emph{Mizan}: we use the evaluation subset of the Mizan corpus~\cite{kashefi2018mizan}, which is acquired based on a manual alignment of famous literary works and their published Persian translations. Overall, the combination of these four high-quality subsets yields an evaluation set that  contains $47k$ sentences, from $4$ different domains (Table~\ref{tab:statistics}.) 

While our main contribution here is providing a more comprehensive \emph{evaluation} of machine translation, we also provide training/dev sets to let the future work create comparable experiments to that of ours.   
We compile our training set at the union of the following datasets: (i) questions obtained from the question paraphrasing task (\S\ref{subsec:qqp}, by translating the QQP instances), (ii) the training set of \emph{Mizan} dataset~\cite{kashefi2018mizan}, (iii) TEP dataset~\cite{pilevar2011tep} and Global Voices dataset~\cite{prokopidis2016parallel}. 
The latter two are not included in our evaluation set due to their noisy translations to prevent any inaccurate evaluations. Note that the \emph{Quran} and \emph{Bible} documents are intentionally not included in the training data, in order to measure models' generalization to unseen documents. 


\section{Experiments}
\label{sec:experiments}
We experiment with several recent LMs, to assess the difficulty of the \parsiglue\ tasks (compared to human expert performance) and also to establish baseline performance of the state-of-the-art mono- and multi-lingual pre-trained models.

All the baseline models used in this work are available online.\footnote{Included in the repository mentioned in footnote~\ref{github-link}.}

\paragraph{Evaluation metrics.}
For each task, we pick a common set of existing metrics: For reading-comprehension, we use \emph{F1} between gold answer and the  response string~\cite{rajpurkar2016squad}; for question paraphrasing, textual entailment, multiple-choice question-answering, and sentiment analysis, we use \emph{accuracy}.   
For the first two sub-tasks of sentiment analysis 
(document-level sentiment, aspect extraction), we use \emph{macro-F1}. 
For the third sub-task (aspect-specific sentiment) we use \emph{accuracy} as our target evaluation metric~\cite{angelidis2018multiple, sun2019utilizing}. For machine translation we use \emph{SacreBLEU}~\cite{post-2018-call}. 

\paragraph{Task splits.}
For each task, we have provided statistics on eval, train, and dev splits in Table~\ref{tab:split_sizes}.
In doing so, we have ensured that enough instances are included in our evaluation sets. 

\paragraph{Human performance.}
To have an estimate of the performance and the difficulty of the challenges, we report human performance on a random subset (100-150) of instances from each task. 
Similar to~\citet{wang2019superglue}, we collect annotations from three human annotators,  adjudicate the inconsistencies and   
evaluate it against the gold labels to estimate human performance for each task. 

\paragraph{Models.}
For evaluation of our baselines, we use state-of-the-art LMs. 
Multilingual BERT (mBERT)~\cite{devlin2019bert} is pre-trained on the masked LM task over 104 languages. 
Additionally, we use two specialized variants of BERT for Persian:  
wikiBERT\footnote{https://github.com/TurkuNLP/wikibert} (trained on Persian Wiki) and ParsBERT~\cite{farahani2020parsbert}.\footnote{
https://github.com/hooshvare/parsbert
}
We also use 
mT5~\cite{xue2020mt5}, which is a multilingual variant of T5~\cite{raffel2020exploring}.

\paragraph{Model selection.}
We train each model with various hyper-parameters and select the best one according to their developement set performance. 
For the BERT-based models, we fine-tune them according to the cross product of the following hyper-parameters: (1) Batch sizes: $\{8, 16\}$ for small/base models and $\{1, 2\}$ for large models; (2) Training epochs: $\{3, 7\}$; (3) Learning-rates: $\{3\times10^{-5}, 5\times10^{-5}\}$. For mT5 models, we fine-tune them for $20k$ steps, dumping checkpoints every $1k$ step.  For the translation task, we trained the models for $200k$ steps since the task has  much larger training data.
We use $10^{-3}$ learning-rate.

\paragraph{Input/output encoding.}
We formulate question paraphrasing (\S\ref{subsec:qqp}) and entailment (\S\ref{subsec:entailment}) tasks as text classification tasks.\footnote{
https://git.io/JYTNr
} 
For sentiment-analysis (\S\ref{subsec:sa}), we follow formulation of~\citet{sun2019utilizing} and encode the instances as questions per aspect. The expected output is the sentiment polarity of the input review with respect to the input aspect-specific question. 
This formulation has the benefit that it is not restricted to a particular domain and its associated set of aspects, unlike alternatives such as multi-class classification.


\paragraph{Experimental setups.}
First, we fine-tune our models on \emph{Persian} (our dataset).
The results of this setup are listed in the top segment of  Table~\ref{tab:main:results:table}.

Following the recent works on generalization across languages~\cite{artetxe2019cross}, we evaluate \emph{English} models on our Persian benchmark. 
We use the commonly used English datasets to supervise mT5 on each task and evaluate the resulting model on the evaluation section of \parsiglue{}. 
The English datasets used here are as follows: 
SQuAD 1.1~\cite{rajpurkar2016squad} for reading comprehension (size: $88k$), 
the union of ARC~\cite{clark2018think}, OpenBookQA~\cite{mihaylov2018can} and CommonsenseQA~\cite{talmor2019commonsenseqa} for multiple-choice question-answering (size: $18k$), 
SNLI~\cite{bowman2015large} for textual entailment (size: $550k$), 
QQP\footnote{See footnote~\ref{ft:qqp}.} for question paraphrasing (size: $350k$),
and Arabic-English subset of OPUS-100~\cite{zhang2020improving} for machine translation (size: $1m$). 
We don't do such mixing for sentiment analysis since existing English datasets are not quite compatible with our sentiment schema.
The results are reported in the middle section of Table~\ref{tab:main:results:table}. 

Finally, we train models on the union of Persian and English datasets. 
Since English datasets tend to be much bigger than Persian ones, we make sure that the batches of training data, on average, contain the same number of instances from each language. Similar treatments of task mixing have also been adopted by \citet{khashabi2020unifiedqa,raffel2020exploring}. The results of this setup are at the bottom segment of Table~\ref{tab:main:results:table}.

\subsection{Results}
\label{sec:results}
Below are key insights from the empirical work: 

\paragraph{Humans do well on \parsiglue.}
As shown in the last row of Table~\ref{tab:main:results:table}, the human upper-bound scores are relatively high across the board. 
This is an indication of a reasonable degree of consensus between the ground-truth and judgments of native speakers and hence, the quality of our dataset. 

\paragraph{Models haven't solved \parsiglue\ yet.}
The majority of the models significantly lag behind human performance.
This is especially true for the mid-sized (`large' or smaller)
models that are commonly used. 
It is encouraging that our largest model (mT5-XL) achieves close to human performance, for certain tasks (e.g., question paraphrasing), however, this model is prohibitively large and it requires a massive amount of compute. 
However, even these large models still struggle for most of the remaining tasks, particularly multiple-choice QA. 


\paragraph{English models successfully transfer to Persian.}
Consistent with the prior observations~\cite{artetxe2019cross}, multilingual models (mT5, in this case) trained with English data show a surprising degree of generalization to other languages (to Persian, in our case).
Training on English data is particularly helpful for challenges that were originally translated from English datasets (such as `qqp` and `mnli`). 

\paragraph{Joint training on English and Persian helps.}
For most of the tasks, combining Persian and English yields better results than training solely on Persian or English data. 

\changed{
 While joint training generally helps, such combinations are not guaranteed to lead to positive gains all the times. 
Whether the ``Eng + Per'' models will beat either of the Persian-only or English-only models depends on whether their strengths (large size of ``Eng'' and distributional alignment of ``Per'') align or go against each other. 
Because of this issue, the combined models are not always better than the individual models. 
}


\section{Discussion}
We now discuss several limitations of the current dataset and the experiments. We then outline several directions for future work.

\paragraph{Beyond current models.}
As shown 
\changed{in the earlier experiments, for most of the tasks}
the current mid-sized models perform significantly worse than humans. 
\changed{
This is particularly pronounced for the multiple-choice QA task where there is over 40\% gap between the model and human performance, and increasing the model size (number of parameters) shows minimal benefits. 
}

\changed{
We hypothesize that the difficulty of our multiple-choice questions (and other tasks, to some extent) for the models are partly due to the reasoning and abstraction needed to answer them. 
For example, the `literature' questions often demand creating connection several pieces of poetry, based on abstract interpretations of their meanings. 
Likewise, most of the `math \& logic' questions require several `hops' of algebraic operations to get to the final answer. 
We hypothesizes these challenges (multi-hop reasoning over high-level abstractions of language) cannot be solely be addressed with more training data. 
and likely require a dramatic rethinking of our architectures design. 
}
For example, the poor performance on `math \& logic' questions might be due to models' inability to comprehend Persian numbers and do logical reasoning with them, a topic that is briefly studied in English~\cite{geva2020injecting}. 
\changed{
There might also be value in exploring multi-task setups across our various tasks ~\cite{zaremoodi2018adaptive}, which we delegate to the future work. 
We hope this benchmark will encourage more of such studies, especially in the context of the Persian language.} 


\paragraph{Coverage of dialects.}
There are other dialects of Persian, including Dari and Tajiki dialects, that are not covered by our dataset. 
We acknowledge this limitation and hope the future work will create broader and more inclusive collections. 


\section{Conclusion}
This work introduced \parsiglue{}, a benchmark for high-level language understanding tasks in Persian. 
We present a careful set of steps we have followed to construct each of the tasks with the help of native speakers (\S\ref{subsec:constructing:tasks}). 
We have presented human scores to establish estimated upper-bounds for each task. This is followed by evaluating state-of-the-art models on each task and quantifying the human-machine gap (\S\ref{sec:experiments}). 

To the best of our knowledge, this is the first work that publishes a language understanding benchmark for Persian language. 
We hope that \parsiglue{} inspires more activity in the Persian NLU tasks, as well as contributing to the latest efforts in multilingual NLU.

\section*{Acknowledgement}
The authors would like to thank Alireza Nourian for providing the OCR system used in the work and the anonymous reviewers for their constructive feedback. 
Thanks to Google's TensorFlow Research Cloud (TFRC) for making research TPUs available. 

\bibliography{ref}
\bibliographystyle{acl_natbib}


\clearpage

\appendix

\end{document}